\documentclass[10pt,letterpaper,journal,compsoc]{IEEEtran}
\ifdefined\pdfobjcompresslevel
\fi

\usepackage{amsmath,amssymb,amsfonts}
\usepackage{mathtools}
\usepackage{amsthm} 
\usepackage{graphicx}
\usepackage{booktabs}
\usepackage{multirow}
\usepackage{array}
\usepackage{xcolor}
\usepackage{algorithm}
\usepackage{algpseudocode}
\algrenewcommand\algorithmicrequire{\textbf{Input:}}
\usepackage{url}
\usepackage{cite}
\usepackage{microtype}
\usepackage[hidelinks]{hyperref}
\usepackage[capitalize,noabbrev]{cleveref}
\makeatletter
\renewcommand{\theHALG@line}{\thealgorithm.\arabic{ALG@line}}
\makeatother
\hypersetup{
  pdftitle={Unified Trajectory Matching Policy Optimization: Diverse T2I Generation and VLA Generalization},
  pdfauthor={Zhiyuan Ma, Jiaming Li, Lingzhen Li, Yu Liu, Xuekai Zhu, Dingkang Liang, Kaiyan Zhang, Jianjun Li, Bowen Zhou, Xiang Bai},
  pdfsubject={Unified trajectory matching for diverse text-to-image generation and vision-language-action generalization},
  pdfkeywords={reinforcement learning post-training, trajectory distribution matching, diffusion policies, flow-matching policies, text-to-image generation, vision-language-action models, generative diversity, out-of-distribution generalization}
}

\newcommand{\unitmpo}{Uni-TMPO}
\newcommand{\paperfigure}[3]{\includegraphics[width=\linewidth]{#1}}
\DeclareMathOperator{\KL}{KL}
\DeclareMathOperator{\softmax}{softmax}

\crefname{section}{Sec.}{Secs.}
\Crefname{section}{Section}{Sections}
\crefname{equation}{Eq.}{Eqs.}
\crefname{figure}{Fig.}{Figs.}
\crefname{table}{Table}{Tables}
\crefname{algorithm}{Alg.}{Algs.}

\begin{document}
\bstctlcite{IEEEtran:BSTcontrol}

\title{Unified Trajectory Matching Policy Optimization:\\
Diverse T2I Generation and VLA Generalization}

\author{Zhiyuan Ma,~\IEEEmembership{Member,~IEEE}, Jiaming Li, Lingzhen Li, Yu Liu, Xuekai Zhu, Dingkang Liang, Kaiyan Zhang, \\
Jianjun Li, Bowen Zhou,~\IEEEmembership{Fellow,~IEEE} and Xiang Bai,~\IEEEmembership{Fellow,~IEEE}
\IEEEcompsocitemizethanks{
\IEEEcompsocthanksitem Z. Ma, J. Li, L. Li, D. Liang and J. Li are with the School of Computer Science and Technology, Huazhong University of Science and Technology, Wuhan, China. (e-mail: \{mzyth,dkliang,jianjunli\}@hust.edu.cn)
\IEEEcompsocthanksitem Yu Liu is with the Institute of Information Engineering, Chinese Academy of Sciences, Beijing, China. (e-mail: liuyu@iie.ac.cn)
\IEEEcompsocthanksitem X. Zhu is with the School of Computer Science and Technology, Shanghai Jiao Tong University, Shanghai, China. (e-mail: xuekaizhu0@gmail.com)
\IEEEcompsocthanksitem K. Zhang is with Frontis.AI, Beijing, China (e-mail: zhangkaiyan@\allowbreak frontis.cn).
\IEEEcompsocthanksitem B. Zhou is with the Department of Electronic Engineering, Tsinghua University, Beijing, China. (e-mail: zhoubowen@tsinghua.edu.cn)
\IEEEcompsocthanksitem X. Bai is with the School of Software Engineering, Huazhong University of Science and Technology, Wuhan, China. X. Bai is the corresponding
author (e-mail: xbai@hust.edu.cn).
}
}

\markboth{IEEE Transactions on Pattern Analysis and Machine Intelligence}%
{Unified Trajectory Matching Policy Optimization}

\IEEEtitleabstractindextext{%
\begin{abstract}
Reward-maximizing reinforcement learning (RL) is widely used to post-train stochastic diffusion and flow policies for text-to-image (T2I) generation. However, reward-maximizing RL causes policy mode collapse even under reference KL or entropy regularization, reducing the policy to a single high-reward mode. In T2I, this produces similar images and reward hacking. When extended to vision-language-action (VLA) models, the same collapse removes alternative successful strategies and weakens task and scene generalization. To address this limitation, we introduce Unified Trajectory Matching Policy Optimization (\unitmpo{}), a unified RL post-training framework for diffusion and flow policies. First, \unitmpo{} converts standardized rewards into a target distribution within each trajectory group and derives the policy distribution from trajectory log probabilities. Then, forward Kullback--Leibler optimization matches the two distributions instead of maximizing expected reward. A progress-conditioned coarse-to-fine scheduler efficiently constructs T2I trajectories. Within the unified framework, feedback-conditioned sampling uses updated observations to construct VLA trajectories. Extensive experiments show that \unitmpo{} achieves higher T2I rewards and VLA ID success rates than the strongest baselines. More importantly, it achieves the best T2I reward--diversity--efficiency trade-off and VLA generalization to held-out tasks and scenes, while real-robot evaluation demonstrates the value of multiple action strategies when the higher-reward target is blocked.
\end{abstract}

}

\maketitle
\IEEEdisplaynontitleabstractindextext

\section{Introduction}\label{sec:intro}

\IEEEPARstart{R}{eward}-maximizing RL has become a common post-training paradigm for diffusion and flow policies. It has been widely studied in T2I generation, where it improves image composition, text rendering, and human preference \cite{black2024ddpo,fan2023dpok,liu2025flowgrpo}. More recently, the same paradigm has been extended to VLA action generation, where online RL improves task completion beyond imitation learning~\cite{chen2025pirl,li2025simplevlarl}. Both applications start from Gaussian noise but sample trajectories differently. T2I produces one image through a complete denoising process, whereas VLA generates multiple action chunks as observations change. Although the sampling procedures differ, both policies assign probabilities to complete trajectories through stochastic diffusion or flow transitions. Each trajectory then receives a reward based on the final image or manipulation result. This shared structure provides a common basis for unified RL post-training and raises a central question: \emph{How can we achieve RL post-training for image and action generation within a unified framework while preserving diverse solution modes and supporting OOD generalization?}

\begin{figure*}[!t]
  \centering
  \paperfigure{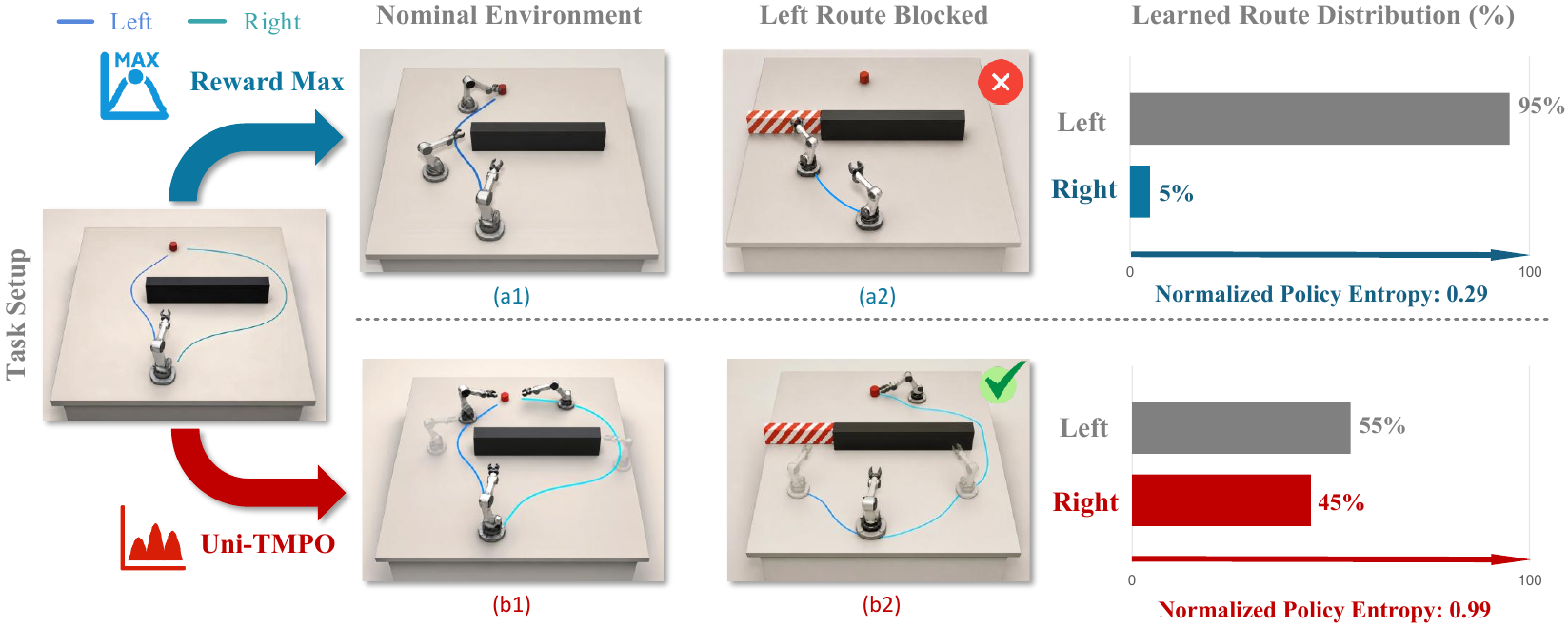}{4.20in}{%
    \textbf{WallDetour strategy coverage and route blocking}\\[3pt]
    Setup: short LEFT and long RIGHT routes around an asymmetric obstacle\\[2pt]
    (a1,b1) ID rollouts after Reward-Max / Uni-TMPO training \quad
    (a2,b2) Zero-adaptation Block-Left evaluation\\[2pt]
    Right: learned LEFT/RIGHT policy mass and normalized route entropy}
  \caption{\textbf{WallDetour experiment.} LEFT is shorter than RIGHT, although both reach the target under the same $0.9\times\mathrm{Success}+0.1\times\mathrm{PathEfficiency}$ reward. Reward Maximization concentrates successful rollouts on LEFT ($0.95/0.05$), whereas \unitmpo{} retains both routes ($0.55/0.45$), with normalized binary route entropies of $0.29$ and $0.99$. When LEFT is blocked at evaluation, only \unitmpo{} reaches the target via RIGHT.}
  \label{fig:walldetour}
\end{figure*}

A central obstacle is \emph{policy mode collapse} under reward-maximizing post-training, where the policy collapses to a single high-reward mode. Sampling images or actions from independent Gaussian noise appears to introduce randomness, but it does not preserve distinct modes after reward-maximizing post-training. As shown in \cref{fig:t2i-qualitative,fig:walldetour}, the resulting policy still maps different noise initializations to similar final images or the same dominant action strategy. In T2I, different pixel-noise samples can produce images with similar object appearances, aesthetic styles, or spatial layouts. Proxy rewards may also rise even as perceptual quality or diversity declines, leading to reward hacking. The recent extension to VLA inherits the same problem. Different action-noise samples can produce one dominant manipulation strategy. As a result, alternative successful routes or action sequences disappear, weakening task and scene generalization. Reference KL and transition entropy can regularize post-training, but they do not consistently retain distinct complete outputs or action strategies. Although explicit diversity rewards directly target this problem, they require domain-specific evaluators~\cite{miao2024diversityrl,he2025gardo}. A unified framework should therefore use a shared optimization objective to preserve multiple successful trajectories while supporting the different trajectory structures of T2I and VLA.

\begin{figure*}[t]
  \centering
  \paperfigure{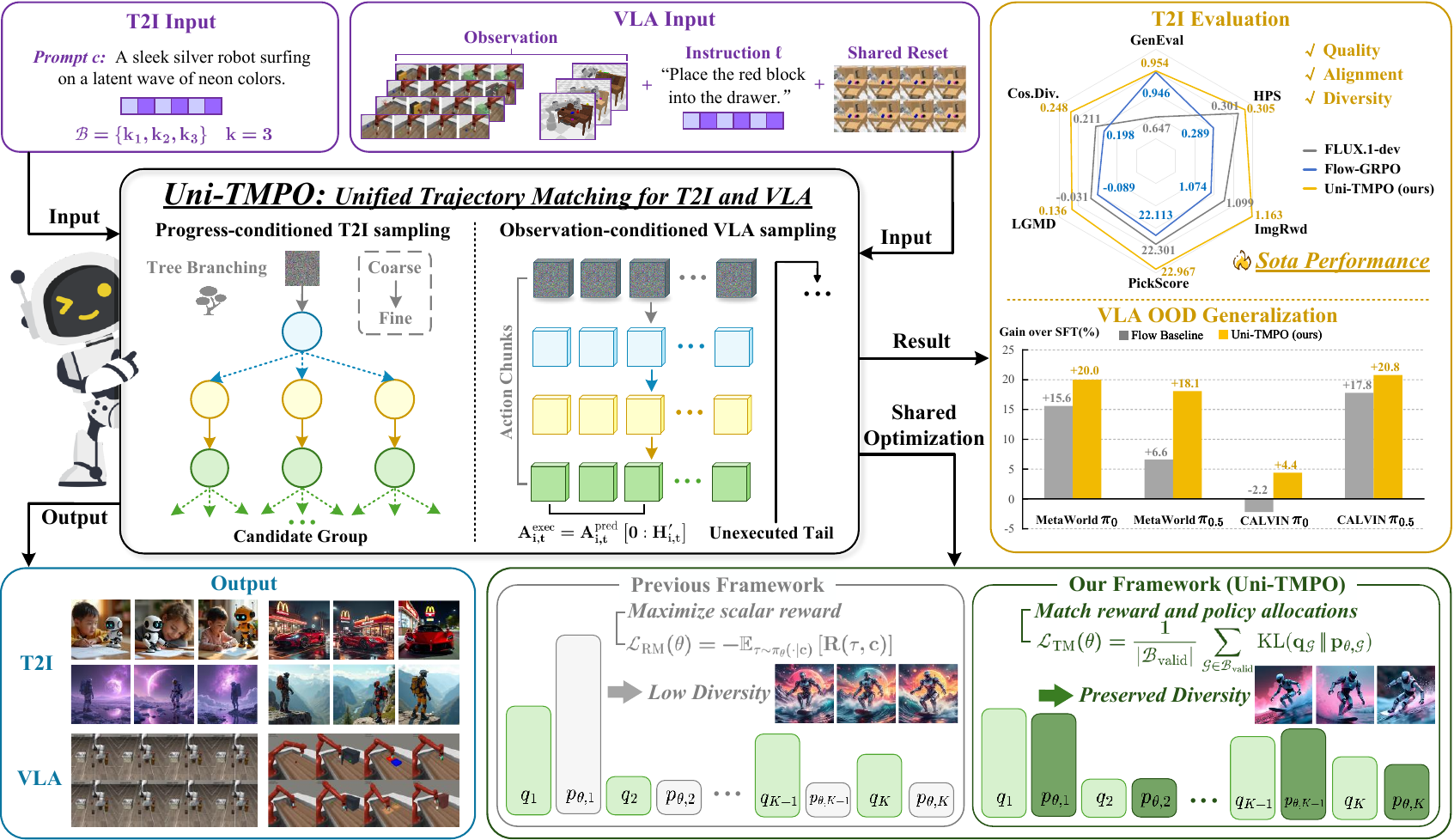}{3.95in}{%
    \textbf{Fig. 1: Uni-TMPO across generation and manipulation}\\[3pt]
    Left: Gaussian image noise, denoising trajectories, and diverse robot images\\[2pt]
    Right: Gaussian action noise and multi-chunk manipulation trajectories\\[2pt]
    Bottom: target distribution $q$, policy distribution $p_\theta$, and trajectory matching}
  \caption{\textbf{Overview of \unitmpo{} for T2I generation and VLA manipulation.} Progress-conditioned coarse-to-fine sampling constructs T2I trajectory groups for each prompt, while feedback-conditioned action-chunk sampling constructs VLA trajectory groups from a shared initialization. Unlike scalar reward maximization, \unitmpo{} matches reward-induced and policy-induced allocations within each group, preserving diverse image modes and action strategies. The right panels summarize T2I performance and VLA OOD generalization.}
  \label{fig:teaser}
\end{figure*}

To answer this question, we introduce \textbf{Unified Trajectory Matching Policy Optimization (\unitmpo{})}, a unified RL post-training framework for stochastic diffusion and flow policies, as shown in \cref{fig:teaser}. For both T2I and VLA, the framework takes a context-conditioned trajectory group, one reward per trajectory, and the corresponding stochastic transition log probabilities as its training inputs. \unitmpo{} replaces expected-reward maximization with trajectory distribution matching within each group sampled under the same context. Standardized rewards form the groupwise Boltzmann target distribution $q$, while accumulated transition log probabilities form the policy distribution $p_\theta$. Forward KL then moves $p_\theta$ toward $q$ and is evaluated exactly over the sampled trajectories. The target can assign substantial probability to multiple high-reward trajectories instead of selecting only the top-ranked sample. Matching this target raises the probability of high-reward trajectories while retaining diverse image modes and alternative action strategies represented in the group. With this objective fixed, trajectory construction is adapted to each application. The T2I implementation collects complete denoising trajectories with \textbf{progress-conditioned coarse-to-fine sampling}. It places branches at earlier denoising steps during the early training stage to explore global structures, then moves them to later steps to refine local details with less redundant denoising. Within the unified framework, \textbf{feedback-conditioned action-chunk VLA sampling} builds trajectories through repeated policy queries. At each query, the policy generates a chunk from the current observation and language instruction; executing its prefix yields the observation for the next query. The transition log probabilities of all chunks jointly determine the trajectory's relative probability within the group. Together, the shared objective and domain-specific trajectory construction provide a unified framework for image and action generation.

The unified framework and its experimental findings are summarized as follows:
\begin{itemize}
  \item \textbf{Unified RL post-training framework.} We introduce a unified trajectory-matching framework for RL post-training of stochastic diffusion and flow policies. It uses groupwise forward KL to match the reward-induced and policy-induced distributions over complete trajectories. We apply this framework to T2I and VLA, with sampling adapted to each domain.
  \item \textbf{Progress-conditioned coarse-to-fine T2I sampling.} The scheduler explores global structures early and refines local details later. Under the same trajectory-matching objective and settings, it achieves higher reward and diversity than the TreeGRPO scheduler while reducing iteration time by $10.2\%$.
  \item \textbf{Feedback-conditioned action-chunk VLA sampling.} Each trajectory comprises action chunks generated under changing observations and a shared language instruction. Their transition log probabilities provide trajectory probabilities for the unified framework.
  \item \textbf{Extensive evaluation.} Under single-reward and joint-reward T2I optimization, \unitmpo{} achieves the best reward--diversity--efficiency trade-off, with LGMD up to $28.4\%$ higher than the strongest baseline. It also achieves the best VLA ID results and exceeds the strongest reward-maximization baselines by an average of $8.0$ percentage points on task OOD generalization and $4.8$ points on scene OOD generalization. Real-robot evaluation further shows how multiple action strategies support task completion when the higher-reward target becomes unavailable.
\end{itemize}

\section{Related Work}\label{sec:related}

\subsection{T2I Post-Training and Generative Diversity}

Diffusion and flow samplers expose sequential stochastic transitions that support policy optimization. DDPO formulates image denoising as a finite-horizon decision process. DPOK adds a reference-policy constraint, while Flow-GRPO applies online RL with group-relative rewards~\cite{black2024ddpo,fan2023dpok,liu2025flowgrpo}. Later work improves exploration and stability through tree collection, regulated clipping, replay, and reward regularization~\cite{ding2026treegrpo,fu2025dynamictreerpo,wang2025grpoguard,zhang2026opgrpo,he2025gardo}. DRaFT backpropagates differentiable rewards through denoising. Diffusion-DPO and D3PO instead learn from pairwise preferences~\cite{clark2024draft,wallace2024diffusiondpo,yang2024d3po}. These methods improve reward but often collapse the policy onto a narrow set of high-reward image modes, reducing generative diversity.

Beyond mode collapse, continued optimization can increase proxy rewards even as perceptual quality and diversity decline, leading to reward hacking~\cite{pan2022reward,skalse2022reward,gao2023scalinglaws}. T2I studies report this behavior together with the trade-off between image quality and diversity~\cite{wu2025rewarddance,wang2025grpoguard,he2025gardo,hong2026rewardhacking}. To improve diversity, existing methods evaluate multiple images generated from the same prompt using a predefined reference distribution~\cite{miao2024diversityrl}. \unitmpo{} instead constructs the groupwise target distribution directly from the scalar rewards already used for conventional RL post-training.

\subsection{VLA Post-Training and OOD Generalization}

Many manipulation tasks admit multiple successful action sequences. ACT predicts temporally coherent action chunks, and Diffusion Policy models conditional multimodal action distributions through denoising~\cite{zhao2023act,chi2023diffusionpolicy}. VLA models such as RT-1, RT-2, Open X-Embodiment, Octo, OpenVLA, $\pi_0$, and $\pi_{0.5}$ condition policies on vision and language~\cite{brohan2022rt1,brohan2023rt2,openx2024rtx,team2024octo,kim2024openvla,black2024pi0,physicalintelligence2025pi05}.

RL post-training of stochastic generative policies has more recently been extended to VLA. GRAPE uses trajectory preferences, ConRFT combines offline values with online updates to consistency policies, and SimpleVLA-RL scales interaction across manipulation tasks~\cite{zhang2024grape,chen2025conrft,li2025simplevlarl}. $\pi_{\mathrm{RL}}$ makes the action-flow policies of $\pi_0$ and $\pi_{0.5}$ stochastic and provides tractable transition log probabilities for on-policy optimization~\cite{chen2025pirl}. \unitmpo{} uses the same policy formulation but matches complete trajectories composed of multiple action chunks. We therefore use $\pi_{\mathrm{RL}}$ as the corresponding reward-maximization baseline. Recent work shows that RL can improve task success while reducing a generative robot policy to one behavior~\cite{longhini2026behavioral}. This loss of alternative strategies motivates trajectory matching for VLA.

Losing alternative action strategies is especially harmful under OOD conditions, because the behavior favored during RL may fail when the task or scene changes. Accordingly, VLA generalization is evaluated using held-out tasks, scene transfer, and controlled execution changes. Online RL can improve generalization under some shifts, although visual robustness remains challenging~\cite{liu2025whatcanrl}. MetaWorld and CALVIN test held-out tasks and scene transfer, respectively~\cite{yu2020metaworld,mees2022calvin}. Controlled benchmarks such as the Colosseum further test policy responses when a previously successful behavior becomes invalid~\cite{pumacay2024colosseum}. Together, these settings motivate our evaluation of task and scene OOD generalization.

\subsection{Trajectory, Flow, and Distribution Matching}

KL-constrained policy optimization has a long history. REPS derives exponential reward weighting under a relative-entropy constraint, and MPO alternates between a reward-weighted nonparametric target and a parametric projection~\cite{peters2010reps,abdolmaleki2018mpo}. PPO optimizes a clipped likelihood-ratio surrogate, while GRPO replaces a learned value baseline with statistics from trajectories sampled under the same context~\cite{schulman2017ppo,shao2024deepseekmath}. These objectives support reward optimization or constrained policy improvement, but they do not directly match the probabilities of complete trajectories.

In contrast, GFlowNets sample complete trajectories in proportion to reward, while Trajectory Balance enforces consistency over complete paths~\cite{bengio2021gflownet,malkin2022tb}. FlowRL applies Boltzmann reward matching to LLM reasoning and learns a prompt-conditioned normalizer $Z_\phi$~\cite{zhu2026flowrl}. \unitmpo{} instead constructs the reward-induced and policy distributions within each group, removing the learned normalizer.

For diffusion models, DAG and Diffusion Generative Flow Samplers (DGFS) fit state-dependent flows with detailed-balance or subtrajectory constraints along a diffusion chain~\cite{zhang2025gflowt2i,zhang2024dgfs}. Gradient-informed variants add local signals~\cite{liu2025gradgflownet}. \unitmpo{} instead matches complete-trajectory probabilities directly to a target derived from terminal rewards. The forward KL is computed exactly over the two groupwise distributions.

For efficient T2I rollout collection, TreeGRPO combines tree sampling with a training-independent random-window schedule and group-relative reward maximization~\cite{ding2026treegrpo}. TMPO~\cite{tmpo2026} introduced a Softmax Trajectory Balance method for T2I post-training. In contrast, \unitmpo{} introduces a unified RL post-training framework based on direct groupwise forward-KL optimization for stochastic diffusion and flow policies across T2I generation and VLA control. Both domains share the optimization objective while using domain-specific trajectory construction for T2I denoising trajectories and VLA action-chunk trajectories; see \cref{sec:domain-scoring}.

\section{Preliminaries}\label{sec:prelim}

\subsection{Stochastic Diffusion and Flow Trajectories}

To define a common post-training interface, we represent both T2I and VLA outputs as context-conditioned stochastic trajectories. Let $c$ denote a conditioning context and $\tau$ a trajectory generated by a stochastic diffusion or flow policy. In image generation, $c$ is a text prompt. One trajectory records the complete denoising rollout from Gaussian image noise to one image. In VLA, the policy is queried again whenever a new observation is available. Each query maps Gaussian action noise to an action chunk, and a prefix is executed before the next query. The environment influences later observations and rewards. Only stochastic policy decisions contribute to the differentiable trajectory score. Although their raw inputs and outputs differ, both domains provide a conditioning context, a complete trajectory, an outcome reward, and the log probabilities of stochastic policy transitions.

Group-based post-training samples $K$ trajectories for the same context and compares their rewards~\cite{shao2024deepseekmath,liu2025flowgrpo}. For embodied rollouts, the shared context contains the language instruction and an initialization identifier. The trajectories therefore begin from the same task condition. This identifier defines the comparison group and is not an input to the action policy. Each sampler therefore returns the group representation $\{(\tau_i,R_i,s_{\theta,i})\}_{i=1}^{K}$, where $R_i$ is the trajectory reward and $s_{\theta,i}$ is computed from its stochastic transition log probabilities. This shared representation enables the same groupwise update in both domains; only the construction of $\tau_i$ and $s_{\theta,i}$ differs between T2I and VLA.

\subsection{Reward Maximization with Multiple Valid Trajectories}

Under this common trajectory representation, conventional post-training maximizes expected scalar reward. When several trajectories obtain the same maximal reward, this objective does not distinguish between assigning all probability to one trajectory and distributing probability across them. A single-mode policy can therefore remain optimal even when multiple successful trajectories are available. The formal statement and proof are given in the supplement.

This ambiguity matters when distinct high-reward trajectories correspond to different image modes or physical strategies. Relative advantages improve credit assignment but do not define a target distribution over useful trajectories. A unified framework must therefore support both trajectory structures while specifying which trajectories are preferred and how probability is distributed among them.

\subsection{Boltzmann Reward Distributions}

To define this target distribution from scalar rewards, we use Boltzmann weighting,
\begin{equation}
q_R(\tau\mid c)
\propto \exp\!\bigl(\beta R(\tau,c)\bigr),
\label{eq:boltzmann-background}
\end{equation}
where the inverse temperature $\beta>0$ controls how strongly the distribution favors higher rewards. Small or moderate values retain probability on several useful trajectories. Large values assign most probability to the highest-reward trajectories. Exponential reward weighting also underlies relative entropy and maximum a posteriori policy search~\cite{peters2010reps,abdolmaleki2018mpo}. We apply this principle within each sampled group. \Cref{sec:method} constructs a standardized Boltzmann target over trajectories that share the same context.

\section{Unified Trajectory-Matching Framework}\label{sec:method}

Using the groupwise Boltzmann distribution over complete trajectories, \unitmpo{} provides a unified RL post-training framework for stochastic diffusion and flow policies. Within each group, rewards define the target distribution $q$, while trajectory scores define the policy distribution $p_\theta$. \unitmpo{} matches the two distributions through a groupwise forward-KL objective. T2I and VLA use domain-specific trajectory sampling and score computation, as detailed in \cref{sec:domain-scoring}.

\subsection{Trajectory Groups}\label{sec:method-path-view}

For each context $c$, the policy samples $K$ trajectories,
\begin{equation}
\mathcal G(c)=\{\tau_i\}_{i=1}^{K}.
\label{eq:trajectory-group}
\end{equation}
In T2I, $c$ is a prompt and each $\tau_i$ is a complete image denoising trajectory. In VLA, $c=(\boldsymbol\ell,\rho)$ contains a language instruction $\boldsymbol\ell$ and an initialization identifier $\rho$. Each $\tau_i$ stores the physical rollout and the stochastic action flow decisions that generated its action chunks. Fixing $c$ ensures that all trajectories address the same conditional problem.

Every trajectory provides a terminal or episodic reward $R_i$ and a differentiable trajectory score $s_{\theta,i}=s_\theta(\tau_i\mid c)$. The reward evaluates the image or physical rollout. For continuous diffusion and flow transitions, the score is constructed from recorded transition log densities. Environment transitions are excluded, while intermediate stochastic transitions are optimized through the trajectory-level loss.

\subsection{Reward-Induced Target Distribution}\label{sec:method-target}

We standardize rewards within each group to remove offsets and reduce sensitivity to evaluator scales, obtaining $\widetilde R_i$. We then construct the groupwise Boltzmann target distribution:
\begin{equation}
\boxed{
q_i
=\frac{\exp(\beta\widetilde R_i)}
{\sum_{j=1}^{K}\exp(\beta\widetilde R_j)}
},\qquad q\in\Delta^{K-1}.
\label{eq:reward-target}
\end{equation}
Here, $\Delta^{K-1}=\{x\in\mathbb{R}_{+}^{K}:\sum_{i=1}^{K}x_i=1\}$ denotes the probability simplex over the $K$ trajectories.
Rewards are evaluator scores, not probabilities. \Cref{eq:reward-target} converts their ordering and relative gaps into a target distribution over the group. Higher-reward trajectories receive larger target probabilities, while trajectories with similar rewards receive similar probabilities. The inverse temperature $\beta$ controls the strength of this preference. Large values approach winner-take-all selection over trajectories.

This construction applies to binary task success, continuous image preference, and episodic robot rewards.

\noindent\textbf{Relation to FlowRL.}
As reviewed in \cref{sec:related}, FlowRL learns a prompt-conditioned normalizer $Z_\phi$ for Boltzmann trajectory matching~\cite{zhu2026flowrl}. \unitmpo{} normalizes the reward target and trajectory probabilities within the same group, so the shared normalizer cancels.

\subsection{Groupwise Trajectory Probabilities}\label{sec:method-policy}

We normalize the trajectory scores within each group to obtain the corresponding trajectory probabilities,
\begin{equation}
\boxed{
p_{\theta,i}
=\frac{\exp(s_{\theta,i})}
{\sum_{j=1}^{K}\exp(s_{\theta,j})}
},\qquad p_\theta\in\Delta^{K-1}.
\label{eq:policy-target}
\end{equation}
Here $s_{\theta,i}$ denotes the trajectory score for trajectory $i$, and $p_{\theta,i}$ is its relative probability within $\mathcal G(c)$. Log-softmax provides a stable implementation. Terms shared by every trajectory cancel; terms shared by only a subset still affect their relative probabilities. See \cref{sec:score-image,sec:score-vla} for the T2I and VLA scores.

Within each group, all trajectories share the same context and score definition, so only their relative probabilities are needed. Comparing $p_\theta$ with the target $q$ identifies high-reward trajectories that receive too little policy probability.

\subsection{Forward-KL Distribution Matching}\label{sec:method-projection}

A group with constant reward provides no relative preference. We therefore define
\begin{equation}
M_{\mathcal G}
=\mathbb I\!\left[\max_iR_i-\min_iR_i>\epsilon_R\right]
\label{eq:informative-group-mask}
\end{equation}
and optimize only the valid groups $\mathcal B_{\mathrm{valid}}=\{\mathcal G\in\mathcal B:M_{\mathcal G}=1\}$. The trajectory-matching loss is
\begin{equation}
\boxed{
\mathcal L_{\mathrm{TM}}(\theta)
=\frac{1}{|\mathcal B_{\mathrm{valid}}|}
\sum_{\mathcal G\in\mathcal B_{\mathrm{valid}}}
\KL(q_{\mathcal G}\|p_{\theta,\mathcal G})
}.
\label{eq:forward-kl-loss}
\end{equation}
If a minibatch contains no informative group, it produces no update from rewards. Rewards are detached during an update. For one group,
\begin{equation}
\KL(q\|p_\theta)
=-\sum_{i=1}^{K}q_i\log p_{\theta,i}+C_q,
\label{eq:forward-kl-expanded}
\end{equation}
where $C_q$ is constant with respect to $\theta$. Differentiating the group softmax gives
\begin{equation}
\nabla_\theta\mathcal L_{\mathcal G}
=\sum_{i=1}^{K}(p_{\theta,i}-q_i)
\nabla_\theta s_{\theta,i}.
\label{eq:forward-kl-gradient}
\end{equation}
If $p_{\theta,i}<q_i$, gradient descent increases the relative probability of trajectory $i$; if $p_{\theta,i}>q_i$, it decreases that probability. The coefficients are bounded, sum to zero, and vanish when the distributions agree. Thus, the update corrects each sampled trajectory toward its target probability instead of selecting the group winner or adding an undirected entropy bonus.

\begin{figure*}[t]
  \centering
  \includegraphics[width=\linewidth]{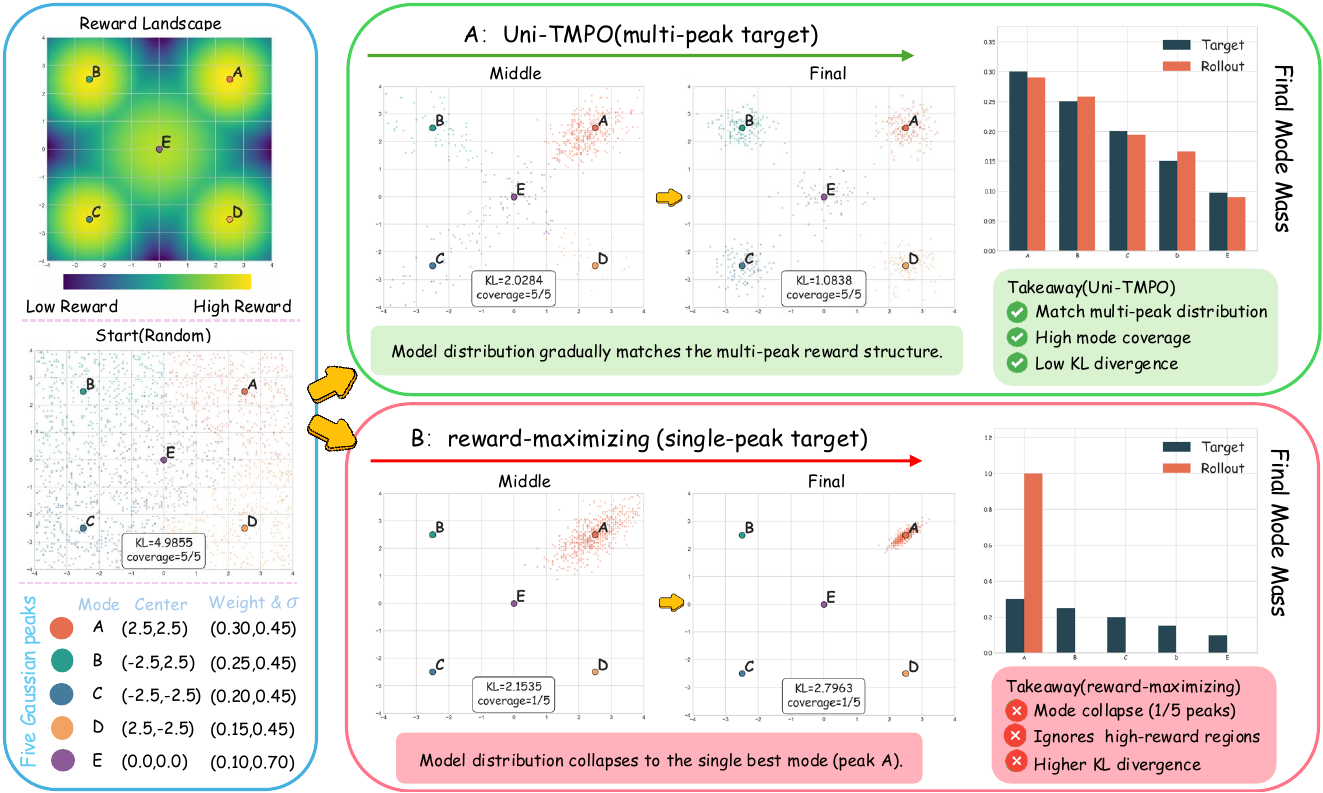}
  \caption{\textbf{Five-mode distribution matching with a three-layer MLP.} The left panels show the five-peak reward landscape and shared initialization. The remaining panels show intermediate and final samples and final mode mass under \unitmpo{} (top) and Reward Maximization (bottom). \unitmpo{} matches the target and preserves all five modes, whereas Reward Maximization collapses to the highest-reward mode.}
  \label{fig:toy-plan}
\end{figure*}

When the trajectory score decomposes as
\begin{equation}
s_{\theta,i}=\sum_{u\in\mathcal U_i}w_{i,u}\ell_{\theta,i,u},
\label{eq:path-score-decomposition}
\end{equation}
the exact transition-level credit is
\begin{equation}
\frac{\partial\mathcal L_{\mathcal G}}
{\partial\ell_{\theta,i,u}}
=w_{i,u}(p_{\theta,i}-q_i).
\label{eq:transition-credit}
\end{equation}
Thus, one correction for trajectory $i$ is propagated through its recorded stochastic decisions. Matching redistributes probability among the sampled trajectories. A mode must first appear in the group before it can receive probability. The toy experiment in \cref{fig:toy-plan} illustrates this behavior.

\subsection{On-Policy Update}\label{sec:method-optimization}

\Cref{alg:unitmpo-update} summarizes the shared \unitmpo{} update.

\begin{algorithm}[H]
\caption{\unitmpo{} on-policy group update.}
\label{alg:unitmpo-update}
\begin{algorithmic}
\Require Policy $\pi_\theta$, contexts $\mathcal C$, group size $K$, temperature $\beta$, threshold $\epsilon_R$
\State $\mathcal L\gets0$, $N\gets0$
\For{$c\in\mathcal C$}
  \State $\mathcal G_c=\{\tau_i\}_{i=1}^{K}\sim\pi_\theta(\cdot\mid c)$
  \State $R_i\gets\operatorname{sg}[R(\tau_i,c)]$ \Comment{detached reward}
  \State $s_{\theta,i}\gets s_\theta(\tau_i\mid c)$ \Comment{trajectory score}
  \If{$\max_iR_i-\min_iR_i>\epsilon_R$}
    \State $q_c\gets\softmax(\beta\widetilde{\mathbf R}_c)$ \Comment{target distribution}
    \State $p_c\gets\softmax(\mathbf s_{\theta,c})$ \Comment{groupwise probabilities}
    \State $\mathcal L\gets\mathcal L+\KL(q_c\|p_c)$, $N\gets N+1$
  \EndIf
\EndFor
\State $\theta\gets\theta-\eta\nabla_\theta(\mathcal L/N)$ if $N>0$
\State \Return $\pi_\theta$
\end{algorithmic}
\end{algorithm}

Rewards and environment transitions remain nondifferentiable. In VLA experiments, gradients pass through the trainable action expert while the vision-language backbone stays frozen. Training uses fresh on-policy groups. A reference-policy KL, when used, remains separate from $\mathcal L_{\mathrm{TM}}$. The supplement provides proofs and implementation details. The next section instantiates trajectory sampling and score computation for T2I and VLA.

\section{Domain-Specific Trajectory Sampling}\label{sec:domain-scoring}

The groupwise forward-KL objective applies to both T2I and VLA, but each domain organizes stochastic transitions differently. We therefore use progress-conditioned coarse-to-fine T2I sampling for complete denoising trajectories and feedback-conditioned action-chunk VLA sampling for robot rollouts. Each procedure also provides the trajectory scores required by the shared objective. Algorithms~\ref{alg:image-collector} and~\ref{alg:vla-collector} summarize the T2I and VLA sampling procedures, respectively.

\begin{algorithm}[H]
\caption{Progress-conditioned coarse-to-fine T2I sampling (shared-prefix branching).}
\label{alg:image-collector}
\begin{algorithmic}
\Require Prompt $c$, policy $p_\theta$, depth $S$, scheduled levels $\mathcal B_\eta$, factor $b$
\State $x_S\sim\mathcal N(0,I)$, $\mathcal F_S\gets\{x_S\}$
\For{$k=S,\ldots,1$}
  \If{$k\in\mathcal B_\eta$}
    \State $\mathcal F_{k-1}\gets\operatorname{Branch}_b(\mathcal F_k;p_\theta,c)$
    \State Record stochastic transitions with log densities in $\zeta$.
  \Else
    \State $\mathcal F_{k-1}\gets\operatorname{Solver}_k(\mathcal F_k)$
  \EndIf
\EndFor
\State $s_{\theta,i}^{\mathrm{img}}\gets
  \sum_{k\in\mathcal K_i^{\mathrm{sto}}}
  \log p_\theta(x_{i,k-1}\mid x_{i,k},c)$
\State \Return $\{(\tau_i,s_{\theta,i}^{\mathrm{img}})\}_{i=1}^{K}$
\end{algorithmic}
\end{algorithm}

\subsection{Progress-Conditioned Coarse-to-Fine T2I Sampling}\label{sec:score-image}

For a text prompt $c$, one generated image corresponds to one complete denoising rollout
\begin{equation}
\tau_i^{\mathrm{img}}=(x_{i,S},\ldots,x_{i,0}),
\qquad x_{i,S}\sim\mathcal N(0,I).
\label{eq:image-trajectory}
\end{equation}
The complete rollout contains several stochastic decisions indexed by $\mathcal K_i^{\mathrm{sto}}$. Given $x_{i,S}$, the image trajectory score is the sum of their transition log densities:
\begin{equation}
s_{\theta,i}^{\mathrm{img}}
=\sum_{k\in\mathcal K_i^{\mathrm{sto}}}
\log p_\theta(x_{i,k-1}\mid x_{i,k},c).
\label{eq:image-path-loglik}
\end{equation}
All leaves use the same number of stochastic transitions, so their scores need no length normalization. The density of their common initial state cancels from the group softmax.

We use shared prefixes to avoid repeating common denoising steps and apply a progress-conditioned coarse-to-fine scheduler to select branch levels~\cite{tmpo2026}. TreeGRPO instead samples one contiguous stochastic window from a training-independent truncated-geometric schedule~\cite{ding2026treegrpo}. Our bounded Beta scheduler places branches earlier during early training to explore composition and global structure. As training proceeds, it moves the branches later to refine appearance and local details. Each leaf records the transition log probabilities along its complete trajectory for the forward-KL objective in \cref{eq:forward-kl-loss}. Shared transitions cancel only when they occur in every trajectory score.

The scheduler changes which complete trajectories enter the group but leaves the matching loss unchanged. It can therefore improve exploration and efficiency without changing how complete trajectories are optimized.

\begin{table*}[!t]
\centering
\caption{Comparison of FLUX.1-dev T2I post-training across compositional image generation, visual text rendering, and human preference alignment. The task-specific training reward is shown in parentheses after each task. Best and second-best results are bolded and underlined.}
\label{tab:diffusion-main}
\setlength{\tabcolsep}{4.5pt}
\resizebox{\textwidth}{!}{
\begin{tabular}{@{}l c cc ccc cc@{}}
\toprule
Method & Time (s)$\downarrow$ & GenEval$\uparrow$ & OCR$\uparrow$ & PickScore$\uparrow$ & HPS$\uparrow$ & ImgRwd$\uparrow$ & LGMD$\uparrow$ & Cos.Div.$\uparrow$\\
\midrule
\multicolumn{9}{c}{\textit{Compositional Image Generation (GenEval)}}\\
\midrule
FLUX.1-dev & -- & 0.647 & -- & \underline{22.301} & \underline{0.301} & 1.099 & $-0.031$ & 0.211\\
DAG-DB & 187.5 & 0.889 & -- & 21.998 & 0.291 & 1.071 & 0.097 & 0.237\\
DGFS-SubTB & 178.6 & 0.917 & -- & 22.210 & 0.298 & \underline{1.107} & \underline{0.113} & \underline{0.241}\\
Flow-GRPO & 160.8 & \underline{0.946} & -- & 22.113 & 0.289 & 1.074 & $-0.089$ & 0.198\\
TreeGRPO & \underline{126.2} & 0.936 & -- & 21.524 & 0.281 & 1.083 & $-0.281$ & 0.184\\
GARDO & 165.1 & 0.926 & -- & 22.261 & 0.292 & 1.087 & 0.009 & 0.235\\
\textbf{\unitmpo{} (ours)} & \textbf{91.9} & \textbf{0.954} & -- & \textbf{22.967} & \textbf{0.305} & \textbf{1.163} & \textbf{0.136} & \textbf{0.248}\\
\midrule
\multicolumn{9}{c}{\textit{Visual Text Rendering (OCR Accuracy)}}\\
\midrule
FLUX.1-dev & -- & -- & 0.591 & 21.968 & \underline{0.292} & \textbf{1.121} & $-0.040$ & 0.215\\
DAG-DB & 135.2 & -- & 0.894 & 21.895 & 0.285 & 1.097 & 0.089 & 0.221\\
DGFS-SubTB & 131.5 & -- & 0.922 & \underline{22.189} & 0.287 & 1.109 & \underline{0.108} & 0.229\\
Flow-GRPO & 121.5 & -- & \underline{0.944} & 21.382 & 0.288 & 1.096 & $-0.089$ & 0.211\\
TreeGRPO & \underline{93.7} & -- & 0.924 & 21.116 & 0.287 & 1.106 & $-0.289$ & 0.183\\
GARDO & 123.6 & -- & 0.934 & 22.104 & 0.288 & 1.089 & 0.061 & \underline{0.231}\\
\textbf{\unitmpo{} (ours)} & \textbf{76.3} & -- & \textbf{0.951} & \textbf{22.281} & \textbf{0.316} & \underline{1.118} & \textbf{0.115} & \textbf{0.235}\\
\midrule
\multicolumn{9}{c}{\textit{Human Preference Alignment (PickScore)}}\\
\midrule
FLUX.1-dev & -- & -- & -- & 22.604 & 0.310 & 1.119 & $-0.056$ & 0.214\\
DAG-DB & 129.0 & -- & -- & 23.691 & 0.338 & 1.504 & 0.139 & 0.228\\
DGFS-SubTB & 124.2 & -- & -- & 23.895 & 0.351 & 1.559 & \underline{0.155} & \underline{0.237}\\
Flow-GRPO & 109.1 & -- & -- & \underline{24.226} & \textbf{0.381} & \underline{1.594} & $-0.104$ & 0.212\\
TreeGRPO & \underline{79.4} & -- & -- & 23.674 & 0.372 & 1.576 & $-0.281$ & 0.179\\
GARDO & 112.5 & -- & -- & 23.976 & 0.347 & 1.566 & $-0.057$ & 0.209\\
\textbf{\unitmpo{} (ours)} & \textbf{68.3} & -- & -- & \textbf{24.301} & \underline{0.380} & \textbf{1.605} & \textbf{0.199} & \textbf{0.258}\\
\bottomrule
\end{tabular}}
\end{table*}

\begin{algorithm}[H]
\caption{Feedback-conditioned action-chunk VLA sampling (sequential replanning).}
\label{alg:vla-collector}
\begin{algorithmic}
\Require Context $(\boldsymbol\ell,\rho)$, policy $\pi_\theta$, group size $K$, replan horizon $H'$, budget $L$
\For{$i=1,\ldots,K$}
  \State $\mathbf o_{i,0}\gets\operatorname{Init}(\rho)$, $t\gets0$
  \While{$\tau_i$ active and $\sum_{v<t}H'_{i,v}<L$}
    \State $(\mathbf A_{i,t}^{\mathrm{pred}},\zeta_{i,t})
      \gets\operatorname{FlowSample}_\theta(\mathbf o_{i,t},\boldsymbol\ell)$
    \State $\mathbf A_{i,t}^{\mathrm{exec}}
      \gets\mathbf A_{i,t}^{\mathrm{pred}}[0{:}H'_{i,t})$
    \State $\mathbf o_{i,t+1}\gets
      \operatorname{EnvStep}(\mathbf o_{i,t},\mathbf A_{i,t}^{\mathrm{exec}})$
    \State $t\gets t+1$ \Comment{execute, observe, and replan}
  \EndWhile
  \State $T_i\gets t$, $R_i\gets\operatorname{Reward}(\tau_i)$
  \State $s_{\theta,i}^{\mathrm{VLA}}\gets
    T_i^{-1}\sum_{t=0}^{T_i-1}\ell_{\theta,i,t}^{\mathrm{chunk}}$
\EndFor
\State \Return $\{(\tau_i,R_i,s_{\theta,i}^{\mathrm{VLA}})\}_{i=1}^{K}$
\end{algorithmic}
\end{algorithm}

\subsection{Feedback-Conditioned Action-Chunk VLA Sampling}\label{sec:score-vla}

For VLA, $c=(\boldsymbol\ell,\rho)$ contains the instruction $\boldsymbol\ell$ and a shared initialization identifier $\rho$. The identifier defines the common initial condition and is not given to the policy.

We use feedback-conditioned action-chunk VLA sampling to construct each trajectory. At valid query $t$, the action expert runs one complete action-flow inference conditioned on the current observation $\mathbf o_{i,t}$ and instruction $\boldsymbol\ell$. It predicts the chunk $\mathbf A^{\mathrm{pred}}_{i,t}=[\mathbf a_{i,t,0},\ldots,\mathbf a_{i,t,H-1}]$. The policy executes the prefix $\mathbf A^{\mathrm{exec}}_{i,t}=\mathbf A^{\mathrm{pred}}_{i,t}[0{:}H'_{i,t})$. The resulting observation conditions the next action chunk.

Because every executed prefix changes the next observation, one action chunk cannot represent the whole rollout. The trajectory score therefore aggregates the action-flow transition log probabilities across policy queries.

Here $H$ is the prediction horizon, $H'$ the replanning horizon, and $T_i$ the number of valid queries. Each executed prefix satisfies $1\leq H'_{i,t}\leq H'\leq H$.

Each chunk maps Gaussian action noise $\mathbf u_{i,t}^{(0)}$ to $\mathbf u_{i,t}^{(S)}$ through stochastic flow transitions indexed by $\mathcal K_{i,t}^{\mathrm{sto}}$. Given the initial noise, observation, and instruction, its score is
\begin{equation}
\ell_{\theta,i,t}^{\mathrm{chunk}}
=\sum_{k\in\mathcal K_{i,t}^{\mathrm{sto}}}
\log p_\theta\!\left(
\mathbf u_{i,t}^{(k+1)}\mid
\mathbf u_{i,t}^{(k)},\mathbf o_{i,t},\boldsymbol\ell
\right).
\label{eq:vla-chunk-logdensity}
\end{equation}
Rewards are evaluated on executed actions. The trajectory score uses the stochastic transitions that generated the full chunks. Environment transitions affect future observations and rewards but have no policy likelihood term.

\begin{figure*}[!t]
  \centering
  \includegraphics[width=0.97\textwidth]{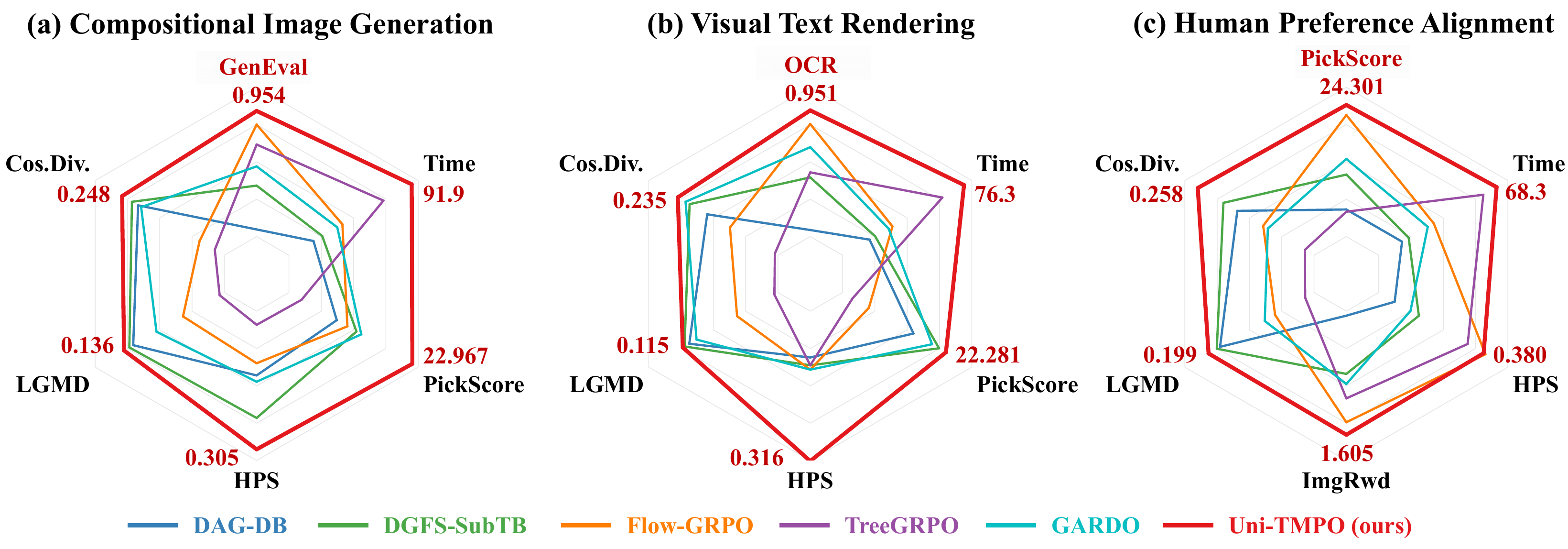}
  \caption{\textbf{T2I reward--diversity--efficiency trade-off.} Metrics are normalized across the post-training methods within each task. Larger values indicate better results on every axis; the Time axis is reversed because shorter iteration time is better. \unitmpo{} provides the strongest overall trade-off across compositional generation, visual text rendering, and human preference alignment.}
  \label{fig:t2i-tradeoff}
\end{figure*}

\begin{figure*}[!t]
  \centering
  \paperfigure{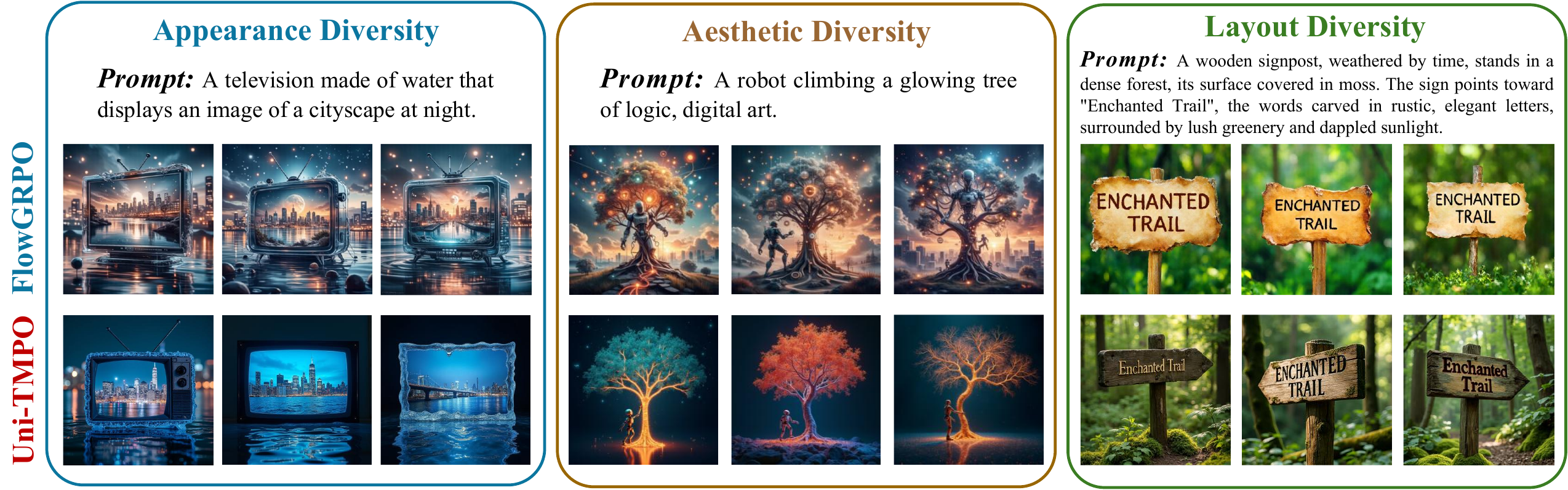}{2.65in}{%
    \textbf{Fig. 4: T2I diversity under matched prompts}\\[3pt]
    Panels: appearance \; aesthetics \; layout\\[2pt]
    Rows: Flow-GRPO \; Uni-TMPO\\[2pt]
    Three stochastic samples per method and prompt}
  \caption{\textbf{T2I diversity under matched prompts.} Each panel compares three stochastic samples per method under the same prompt. Compared with Flow-GRPO, \unitmpo{} produces broader variation in object appearance, aesthetic style, and spatial layout. \Cref{tab:diffusion-main} reports quantitative results.}
  \label{fig:t2i-qualitative}
\end{figure*}

A VLA trajectory contains $T_i$ chunk-generation queries. To compare trajectories with different query counts, we average their chunk scores:
\begin{equation}
s_{\theta,i}^{\mathrm{VLA}}
=\frac{1}{T_i}
\sum_{t=0}^{T_i-1}
\ell_{\theta,i,t}^{\mathrm{chunk}}.
\label{eq:vla-normalized-score}
\end{equation}
This length-normalized trajectory score is the log of the geometric mean of the valid chunk densities. Group softmax converts it into the relative trajectory probabilities used by the matching objective.

Masks exclude padded policy queries, deterministic flow steps, and padded action entries. The full noisy chunk is still supplied to the action expert. Tensor definitions and gradient details are given in the supplement.

Collection begins from a shared initialization. All slots use the same instruction and initial simulator state but receive independent action noise. After execution begins, their observations and action chunks evolve independently. Each trajectory records the physical episode used for reward and the stochastic decisions used for policy scoring.

\subsection{Shared Objective Across Domains}

The T2I and VLA probability calculations produce the same two groupwise distributions required by \unitmpo{}. Forward KL aligns the policy distribution with the reward-induced target in both domains. The following experiments evaluate this unified objective in both domains.

\section{Experiments}\label{sec:exp}

We evaluate the same trajectory-matching objective under both trajectory structures. The T2I study measures reward, diversity, and training efficiency. The VLA study evaluates ID performance and task and scene OOD generalization.

\begin{table*}[!t]
\centering
\caption{In-distribution VLA performance of $\pi_0$ and $\pi_{0.5}$. Baseline configurations follow $\pi_{\mathrm{RL}}$~\cite{chen2025pirl}. LIBERO and MetaWorld-MT50 report average task success, while CALVIN-D reports five-subtask sequence completion (Len-5). All values are percentages. Bold values indicate the best result for each backbone.}
\label{tab:vla-main}
\setlength{\tabcolsep}{4pt}
\resizebox{\textwidth}{!}{%
\begin{tabular}{llccccc cc}
\toprule
Backbone & Method & \multicolumn{5}{c}{LIBERO} & MetaWorld-MT50 & CALVIN-D\\
\cmidrule(lr){3-7}\cmidrule(lr){8-8}\cmidrule(lr){9-9}
& & Spatial & Object & Goal & Long & Avg. & Avg. & Len-5\\
\midrule
\multirow{4}{*}{$\pi_0$}
& SFT & 65.3 & 64.4 & 49.8 & 51.2 & 57.6 & 50.8 & 57.5\\
& Flow-SDE & 98.4 & 99.4 & 96.2 & 90.2 & 96.1 & 78.1 & 61.7\\
& Flow-Noise & 99.0 & 99.2 & 98.2 & \textbf{93.8} & 97.6 & 85.8 & 59.9\\
& \textbf{\unitmpo{} (ours)} & \textbf{99.2} & \textbf{99.6} & \textbf{98.8} & 93.6 & \textbf{97.8} & \textbf{88.6} & \textbf{63.9}\\
\midrule
\multirow{4}{*}{$\pi_{0.5}$}
& SFT & 84.6 & 95.4 & 84.6 & 43.9 & 77.1 & 43.8 & 61.3\\
& Flow-SDE & \textbf{99.6} & \textbf{100.0} & 98.8 & 93.0 & 97.9 & 70.7 & 87.0\\
& Flow-Noise & \textbf{99.6} & \textbf{100.0} & 99.6 & 94.0 & 98.3 & 66.1 & 84.5\\
& \textbf{\unitmpo{} (ours)} & \textbf{99.6} & \textbf{100.0} & \textbf{100.0} & \textbf{94.8} & \textbf{98.6} & \textbf{72.8} & \textbf{89.2}\\
\bottomrule
\end{tabular}}
\end{table*}

\begin{figure*}[!t]
  \centering
  \includegraphics[width=\textwidth]{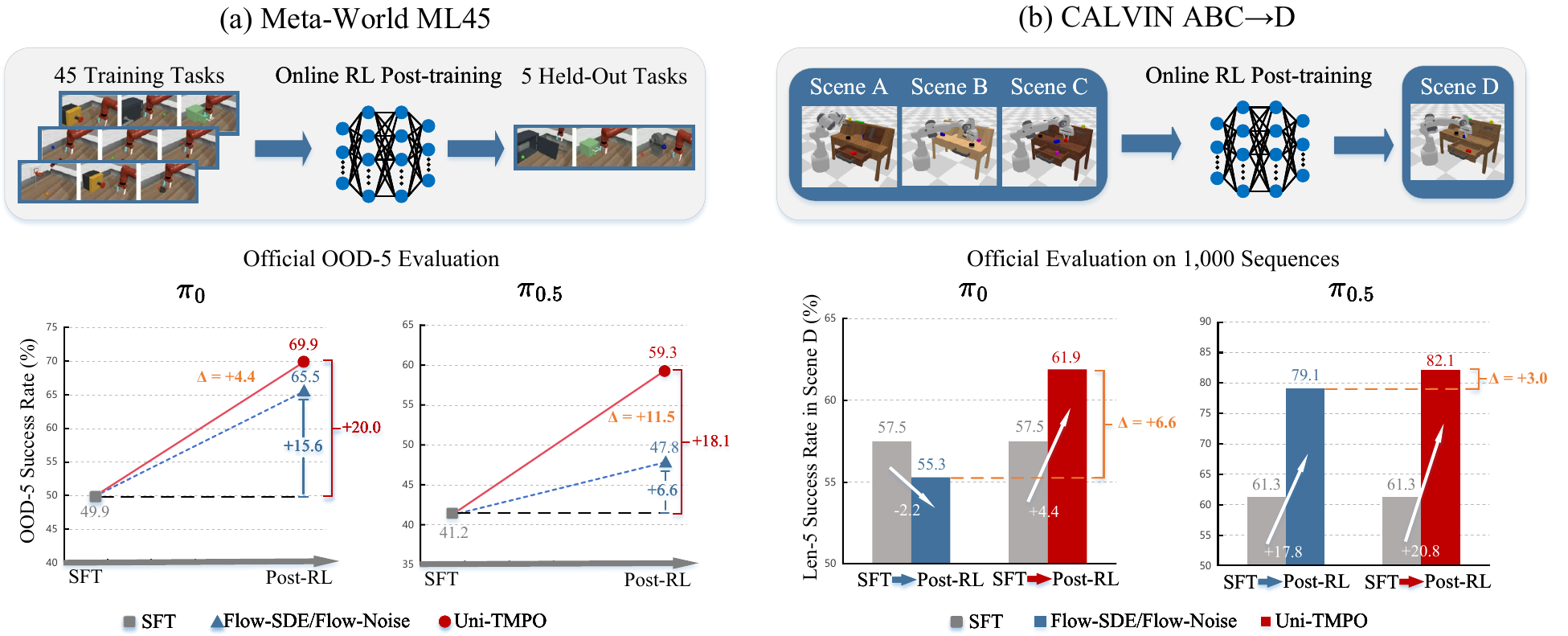}
  \caption{\textbf{Task and scene OOD generalization after VLA post-training.} MetaWorld ML45 uses 45 tasks for online RL and evaluates five unseen tasks. CALVIN uses Scenes ABC for supervised initialization and online RL, then evaluates five-subtask completion (Len-5) in Scene D over 1,000 sequences. The lower panels compare SFT and post-RL performance for $\pi_0$ and $\pi_{0.5}$.}
  \label{fig:vla-ood}
\end{figure*}

\begin{figure*}[!t]
  \centering
  \includegraphics[width=\textwidth]{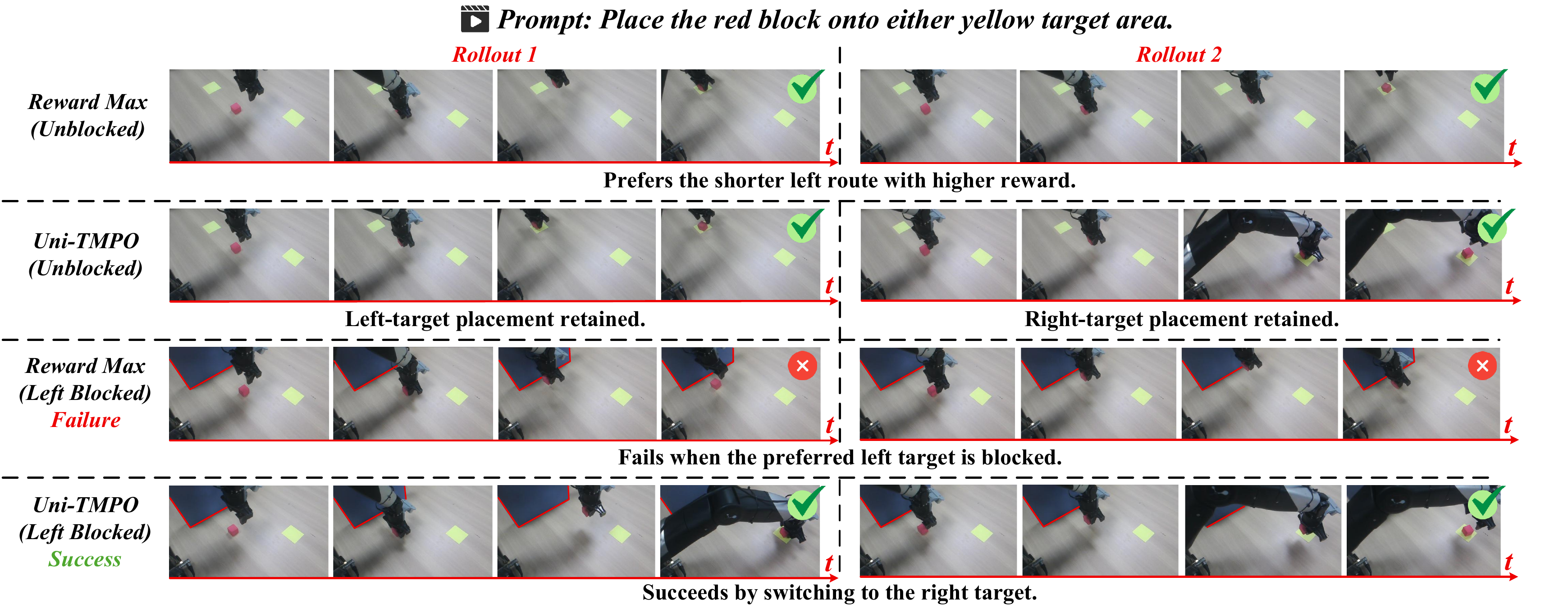}
  \caption{\textbf{Real-robot dual-target placement.} The robot is instructed to place the red block on either yellow target. The left target receives slightly higher reward because it is closer. In the unblocked setting, both Reward Maximization rollouts select the left target, whereas \unitmpo{} succeeds at both targets. After the left target is blocked, Reward Maximization fails, while \unitmpo{} succeeds through the alternative right-target strategy.}
  \label{fig:real-robot}
\end{figure*}

\subsection{T2I Generation}\label{sec:exp-diffusion}

\subsubsection{Experimental Setup}

We use FLUX.1-dev~\cite{flux2024} with LoRA~\cite{hu2022lora}. We consider three main single-reward protocols. GenEval measures compositional correctness~\cite{ghosh2023geneval}. OCR accuracy evaluates visual text rendering~\cite{chen2023textdiffuser}. PickScore measures human preference~\cite{kirstain2023pickapic}. GenEval provides a near-binary correctness score. OCR accuracy rewards exact symbol sequences, whereas PickScore assigns continuous preference scores to individual images. The main comparison is reported in \cref{tab:diffusion-main}. Additional joint-reward experiments combine HPS-v2.1, ImageReward, and PickScore at equal weight or jointly optimize OCR and PickScore~\cite{wu2023hpsv2,xu2023imagereward}. Their full configurations and results are reported in the supplemental material.

For each prompt, we sample a group of $K$ image trajectories with the same depth. The main $K{=}27$ setting uses three stochastic branch levels with branch factor three. Every leaf contributes the same number of transition log densities to \cref{eq:image-path-loglik}, so path length cannot change its relative group probability. \unitmpo{} uses the progress-conditioned coarse-to-fine scheduler in \cref{sec:score-image}, while TreeGRPO retains its random-window scheduler.

\subsubsection{Metrics and Baselines}

Task reward alone cannot diagnose mode collapse. We therefore report two diversity measures on matched prompt groups. LGMD detects near duplicates in VAE latent space~\cite{kingma2014vae}. Cosine diversity measures feature-space diversity using DINOv2 embeddings~\cite{oquab2024dinov2}. The qualitative comparison shows changes in object appearance, aesthetic style, and spatial layout. Within each protocol, all methods share inputs, generation seeds, and inference settings.

The distribution-matching baselines are DAG-DB and DGFS-SubTB~\cite{zhang2025gflowt2i,zhang2024dgfs}. DAG-DB applies one-step detailed balance with a state-dependent flow, while DGFS-SubTB matches diffusion subtrajectories. Both fit intermediate flows along the denoising chain. The reward-maximization baselines are Flow-GRPO~\cite{liu2025flowgrpo} and TreeGRPO~\cite{ding2026treegrpo}; GARDO~\cite{he2025gardo} additionally regularizes diversity. TreeGRPO retains its training-independent random-window schedule, while \unitmpo{} uses the progress-conditioned scheduler in \cref{sec:score-image}. All comparisons use the same prompts, seeds, group size, decoding steps, and number of generated images.

\subsubsection{Main Results}

Flow-GRPO and TreeGRPO improve the optimized task metric but reduce LGMD and cosine diversity relative to FLUX.1-dev in all three protocols. \unitmpo{} instead improves the primary task metric and both diversity measures. It reaches the highest GenEval score of $0.954$ and the highest OCR accuracy of $0.951$. In preference optimization, it obtains the highest ImageReward and PickScore and the second-highest HPS. These results show that the optimized reward can improve without reducing measured visual diversity.

DAG-DB and DGFS-SubTB retain positive LGMD, which supports distribution matching as a way to preserve visual modes. Both methods fit intermediate flows, and DGFS also combines constraints over subtrajectories. In the three-transition FLUX setting, \unitmpo{} directly matches complete trajectories to a target derived from terminal rewards. It achieves higher primary metrics and $42$--$49\%$ lower iteration time than DGFS in this setting.

Across the three protocols, \unitmpo{} provides the best reward--diversity--efficiency trade-off among the compared methods. \Cref{fig:t2i-tradeoff} summarizes the task metrics, diversity measures, and iteration time on a common within-task scale. Training dynamics are reported in the supplemental material.

The supplement evaluates the progress-conditioned coarse-to-fine scheduler against fixed positions and the TreeGRPO scheduler. The \unitmpo{} objective and all other settings remain fixed, so the comparison isolates the effect of the scheduler during T2I post-training.

The category-level GenEval analysis in the supplement shows that the improvement is not confined to one compositional skill. \unitmpo{} is best or tied for best in all six categories, covering object composition, counting, color, position, and attribute binding.

Figure~\ref{fig:t2i-qualitative} compares three forms of diversity under matched prompts. For appearance, \unitmpo{} varies the television shape, screen style, and surrounding scene. For aesthetics, it varies the color palette, lighting, and visual style of the glowing-tree scene. For spatial layout, it varies the sign shape, orientation, viewpoint, and position within the forest scene. Flow-GRPO produces more similar samples in each case. The variations remain consistent with the shared prompt. These examples complement the quantitative diversity metrics reported in \Cref{tab:diffusion-main} across protocols.

\subsubsection{Multi-Reward Evaluation}

We also jointly optimize OCR and PickScore as two distinct reward signals. With weights scheduled from $3{:}1$ to $1{:}3$, \unitmpo{} reaches $0.932$ OCR accuracy and $23.897$ PickScore, with LGMD $0.155$ and cosine diversity $0.248$. Flow-GRPO obtains $0.928$, $23.081$, $-0.049$, and $0.209$, respectively. Fixed $1{:}1$ weighting gives the same ordering. The supplement reports the full comparison and ablations of the coarse-to-fine scheduler, reference KL, branch structure, and $\beta$.

\subsection{VLA Manipulation}\label{sec:exp-vla}

\subsubsection{Training and Evaluation Setup}

The VLA experiments use SFT-initialized OpenPI policies in RLinf~\cite{black2024pi0,chen2025pirl,zang2025rlinf}. We compare \unitmpo{} with the Flow-SDE and Flow-Noise reward-maximization variants of $\pi_{\mathrm{RL}}$. During RL, the vision-language backbone and language model remain frozen. Only the approximately 300M-parameter action expert is optimized. A rollout batch contains 64 environments, forming eight groups of size eight with a shared instruction and initial state. Eight rollout epochs collect 512 trajectories per training iteration. For fairness, all methods use identical environments, resets, interaction budgets, and horizons. The supplement provides training and evaluation details.

The VLA study evaluates ID performance, task and scene OOD generalization, and multiple action strategies in WallDetour and a real-robot blocked-target experiment.

\subsubsection{In-Distribution Performance}

LIBERO~\cite{liu2023libero} contains Spatial, Object, Goal, and Long suites. Each suite has ten tasks and 50 fixed resets per task. MetaWorld-MT50~\cite{yu2020metaworld} evaluates 50 tasks with ten trials each and reports a macro average over four difficulty groups. CALVIN~\cite{mees2022calvin} contains four scenes, A--D. Its ID protocol uses supervised data from Scenes ABC, followed by online RL and evaluation in Scene D (ABC-SFT $\rightarrow$ D-RL $\rightarrow$ D-eval). The metric is completion of a five-subtask sequence. These settings measure ID improvement after RL.

\Cref{tab:vla-main} reports the ID comparison. With $\pi_0$, \unitmpo{} reaches 97.8\% on LIBERO, 88.6\% on MT50, and 63.9\% on CALVIN-D. These results exceed the strongest $\pi_{\mathrm{RL}}$ reward-maximization baseline by 0.2, 2.8, and 2.2 points, respectively. With $\pi_{0.5}$, the corresponding results are 98.6\%, 72.8\%, and 89.2\%. The gains are 0.3, 2.1, and 2.2 points. LIBERO leaves little headroom under this protocol. MT50 and CALVIN-D reveal larger differences through multitask manipulation and long-horizon completion. The next experiments evaluate whether these gains extend to OOD generalization.

\subsubsection{Out-of-Distribution Generalization}\label{sec:exp-vla-ood}

The OOD study evaluates two forms of generalization. MetaWorld ML45 performs online RL on 45 tasks and reports macro success on the five official unseen tasks.

CALVIN measures scene transfer. Supervised initialization and online RL use Scenes ABC, followed by evaluation in Scene D (ABC-SFT $\rightarrow$ ABC-RL $\rightarrow$ D-eval). The official evaluator processes 1,000 five-subtask sequences. We report Len-5, the percentage completing all five subtasks.

Both backbone comparisons include SFT, Flow-SDE, Flow-Noise, and \unitmpo{}. \Cref{fig:vla-ood} compares the SFT and post-RL endpoints for both backbones. Flow-SDE and Flow-Noise overlap because they obtain identical aggregate OOD scores; the connecting segments show endpoint changes rather than training curves over successive iterations.

On MetaWorld, \unitmpo{} reaches 69.9\% with $\pi_0$ and 59.3\% with $\pi_{0.5}$. These results improve over the corresponding SFT checkpoints by 20.0 and 18.1 points and exceed the corresponding $\pi_{\mathrm{RL}}$ reward-maximization baselines by 4.4 and 11.5 points. The consistent gains across both backbones demonstrate that \unitmpo{} generalizes beyond the 45 training tasks to unseen manipulation tasks.

On CALVIN ABC$\rightarrow$D, \unitmpo{} obtains 61.9\% and 82.1\% Len-5 success. With $\pi_0$, the flow baselines fall 2.2 points below SFT; \unitmpo{} improves by 4.4 points and finishes 6.6 points higher. With $\pi_{0.5}$, \unitmpo{} improves over SFT by 20.8 points and exceeds the flow baselines by 3.0 points. Together, MetaWorld and CALVIN show consistent gains under task and scene changes.

\noindent\textbf{WallDetour and real-robot evaluation.}\label{sec:exp-walldetour}
We use WallDetour and a real-robot dual-target placement task to evaluate whether multiple action strategies remain effective after a higher-reward option is blocked.

WallDetour provides two successful routes to the same target. LEFT is shorter and receives slightly higher reward, while RIGHT provides a longer alternative. Reward Maximization and \unitmpo{} start from the same SFT policy trained on both routes and use the same reward without route labels or diversity bonuses. Reward Maximization concentrates successful rollouts on LEFT, with LEFT/RIGHT mass $0.95/0.05$ and normalized route entropy $0.29$. \unitmpo{} retains both routes, reaching $0.55/0.45$ and entropy $0.99$. After LEFT is blocked, Reward Maximization fails, whereas \unitmpo{} completes the task through RIGHT.

We further evaluate the same effect in real-robot dual-target placement, as shown in \cref{fig:real-robot}. The robot is instructed to place the red block on either yellow target. Both placements complete the instruction, but the shorter left placement receives slightly higher reward. In the unblocked setting, the displayed Reward Maximization rollouts both select the left target, whereas \unitmpo{} succeeds at both targets. After the left target is blocked, Reward Maximization fails, while \unitmpo{} completes the task using the right-target strategy.

Together, these experiments show that multiple action strategies support task completion after the higher-reward option becomes unavailable.

\section{Conclusion}\label{sec:conclusion}

Reward-maximizing post-training collapses stochastic diffusion and flow policies onto a narrow set of high-reward trajectories. \unitmpo{} instead matches reward-induced and policy-induced probabilities over trajectories sampled under the same context. For T2I, complete-trajectory matching and the progress-conditioned coarse-to-fine scheduler provide the best reward--diversity--efficiency trade-off among the compared methods. The comparisons with DAG and DGFS further support direct terminal-reward matching for the short stochastic trajectories used in FLUX. The same objective extends to VLA trajectories composed of multiple action chunks. \unitmpo{} achieves the best ID performance and task and scene OOD generalization. WallDetour and the real-robot blocked-target experiment further show that multiple action strategies support task completion when the higher-reward option becomes unavailable. Together, these results demonstrate that groupwise trajectory matching provides a unified RL post-training framework for image and action generation. T2I and VLA use different trajectory sampling procedures but share the same groupwise forward-KL update. Overall, the framework preserves diverse solution modes and supports OOD generalization across stochastic diffusion and flow policies.

Future work may extend the RL post-training framework to video, audio, and 3D generation; molecular and protein design; and diffusion-based planning.

\bibliographystyle{IEEEtran}
\bibliography{references}

\end{document}